\documentclass[sigconf,edbt]{acmart-edbt2023}

\def\BibTeX{{\rm B\kern-.05em{\sc i\kern-.025em b}\kern-.08em
    T\kern-.1667em\lower.7ex\hbox{E}\kern-.125emX}}

\usepackage{booktabs} % For formal tables

\setcopyright{rightsretained}

\acmDOI{}

\acmISBN{978-3-89318-092-9}

\acmConference[EDBT 2023]{26th International Conference on Extending Database Technology (EDBT)}{28th March-31st March, 2023}{Ioannina, Greece}
\acmYear{2023}

\theoremstyle{plain}
\newtheorem{thm}{Theorem}% reset theorem numbering for each chapter
\theoremstyle{definition}
\newtheorem{defn}[thm]{Definition} % definition numbers are dependent on theorem numbers

\usepackage[linesnumbered,ruled,vlined]{algorithm2e}
\usepackage{enumitem}
\usepackage{multirow}
\usepackage{multicol}
\usepackage{makecell}
\usepackage{graphicx}
\usepackage[normalem]{ulem}
\usepackage[font=footnotesize]{subfig}
\usepackage{amsthm}
\usepackage{amsmath}

\begin{document}
\title{POLITE: Pushing Online Learning on Human Activities at the Extreme Edge}
%Hakim suggested this -- feel free to edit or discard please
\title{On Human Physical Activity Recognition and Tracking: \\Handling Catastrophic Forgetting for Incremental Learning on the Edge}

\title{On Handling Catastrophic Forgetting for Incremental Learning of Human Physical Activity on the Edge}

\author{Jingwei Zuo}
\affiliation{Technology Innovation Institute\\
  \city{Abu Dhabi}
  \country{UAE}}
\email{jingwei.zuo@tii.ae}

\author{George Arvanitakis}
\affiliation{Technology Innovation Institute\\
  \city{Abu Dhabi}
  \country{UAE}}
\email{george.arvanitakis@tii.ae}

\author{Hakim Hacid}
\affiliation{Technology Innovation Institute\\
  \city{Abu Dhabi}
  \country{UAE}}
\email{hakim.hacid@tii.ae}

\begin{abstract}
% Problem definition 
Human activity recognition (HAR) has been a classic research problem. In particular, with recent machine learning (ML) techniques, the recognition task has been largely investigated by companies and integrated into their products for customers. 
However, most of them apply a predefined activity set and conduct the learning process on the cloud, hindering specific personalizations from end users (i.e., edge devices). Even though recent progress in Incremental Learning allows learning new-class data on the fly, the learning process is generally conducted on the cloud, requiring constant data exchange between cloud and edge devices, thus leading to data privacy issues.
In this paper, we propose PILOTE, which pushes the incremental learning process to the extreme edge, while providing reliable data privacy and practical utility, e.g., low processing latency, personalization, etc. 
In particular, we consider the practical challenge of extremely limited data during the incremental learning process on edge, where catastrophic forgetting is required to be handled in a practical way. 
We validate PILOTE with extensive experiments on human activity data collected from mobile sensors. The results show PILOTE can work on edge devices with extremely limited resources while providing reliable performance.
\end{abstract}

\maketitle
\section{Introduction}
%JZ: 1. add related work on Activity Recognition and Catastrophic Forgetting; 2. Illustrate the constraints of previous approaches in our context; 

% 1. Human activity recognition 
Human activity recognition (HAR) has gained, in recent years, a great interest from both the research community and industry players. The activity data can be collected from multiple data sources, such as images \cite{vrigkas2015review}, videos \cite{ke2013review}, GPS trajectories \cite{el2022learning}, smart sensors \cite{attal2015physical}, etc. Among which, the activity data collected from wearable sensors are widely studied and applied in real-life products by almost all tech giants, e.g., Huawei-Sussex locomotion datasets \cite{huaweisussex2018}, Google platform~\cite{googlelink}, Samsung health activity trackers~\footnote{Samsung Health: \url{https://www.samsung.com/global/galaxy/apps/samsung-health/} (visited on Sep. 21, 2022)}, and Apple CMMotionActivity~\cite{applelink}. However, these centralized and cloud-based solutions usually lack of flexibility to user's personalization needs. Since data exchanges between cloud and edge devices are usually required, data privacy and processing latency issues may also occur.

\begin{figure}
\centering
\includegraphics[width=0.8\linewidth]{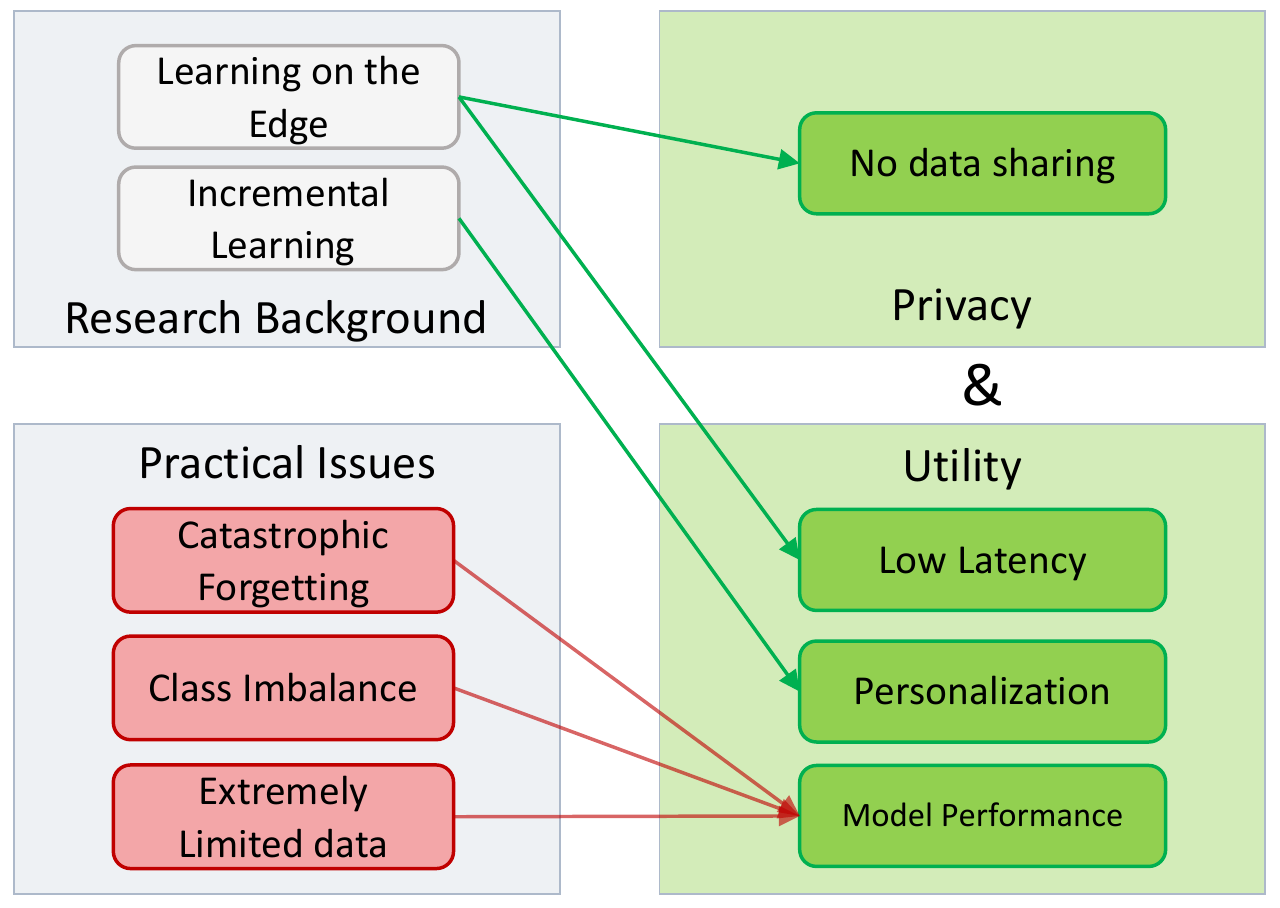}
\caption{Incremental Human Activity Learning on the Edge: a brief context overview of PILOTE.}
\label{fig:research_background}
\vspace{-1em}
\end{figure}

% 2. Edge ML: overview and problematics 
\textsf{Edge} \textsf{ML}~\cite{Lee2018} has brought the processing in \textsf{ML} to the edge of the network and intends to adapt \textsf{AI} technologies to the edge environments. Generally, \textsf{Edge} environment refers to the end-user side pervasive environment composed of devices from both the base station and end device levels.
%can not fulfill the Privacy \& Utility requirements for personalized user experiences, 
As shown in Figure \ref{fig:research_background}, moving ML from a centralized cloud server to edge devices brings benefits, such as reduced communication cost or latency and preserving data privacy. However, it is challenging for the learning process because of the limited computational resources and data availability in each edge device to train a decent ML model~\cite{s22072665}. 
% Other characteristics to be considered while developing/deploying ML models on Edge devices are mobility support, location, proximity and context awareness~\cite{KHAN2019219}, reliability of the service (quality-of-service), trust and explainability~\cite{Murshed_2022}. 

Learning on the Edge exposes new problems that were not necessarily taken care of in the classical ML domain. The most evident issue is undoubtedly the dynamics of the data. In fact, Edge learning assumes that the observed world is (i) partial, i.e., not all the situations are observed, and (ii) dynamic, i.e., new situations may emerge. The assumptions impose new challenges on ML methods to incrementally learn new situations without forgetting previously acquired knowledge. In other words, ML methods on the Edge have to handle the catastrophic forgetting problem~\cite{kirkpatrick2017overcoming}. In the context of human activity recognition, since activity data are generated on the fly but not in a batch mode, class imbalance is also a practical challenge when learning from the dynamic streams. 
%the user's personalization of learning new activities should not deteriorate the model's reliability on previously learned ones.

The problem of catastrophic forgetting has a long history in machine learning and has recently been largely studied in deep learning \cite{de2021continual}. Some of the most recent works have addressed this problem from multiple perspectives, which can be divided into three categories: Replay methods (i.e., store data samples) \cite{rolnick2019experience}, regularization-based methods \cite{rebuffi_icarl_2017} and parameter isolation methods \cite{rusu2016progressive}. However, these proposals are all designed for centralized/cloud-based solutions and not applicable in Edge learning context with limited resources.

% 3. Our proposals and contributions 
In this paper, considering the aforementioned Privacy \& Utility challenges as shown in Figure \ref{fig:research_background}, we propose \textsf{PILOTE}: \textbf{P}ushing \textbf{I}ncremental \textbf{L}earning \textbf{O}n human activities at the ex\textbf{T}reme \textbf{E}dge, a method that contributes to reducing the effect of catastrophic forgetting with a focus on a practical case: incremental human physical activity recognition on the Edge. 
% brief presentation of PILOTE
The Siamese Network-based model \cite{koch2015siamese} with supervised Contrastive Loss \cite{khosla_supervised_2020} allows learning from extremely limited data with class imbalance. A jointly optimized Distillation Loss prevents the model from forgetting the previously learned knowledge by guiding the embedding space's building. 
The performed experiments on our collected human activity data demonstrate that the proposed approach properly handles the addition of new classes while maintaining good performance over the previously learned classes. 

We summarize the paper’s main contributions as follows:
\begin{itemize}
    \item \textbf{Incremental Edge Learning:} We introduce a new primitive, PILOTE, for incremental human activity learning on the Edge, which is rarely investigated by the community.
    \item \textbf{Catastrophic Forgetting on Edge Learning:} We study and tackle the catastrophic forgetting problem in PILOTE when learning on the Edge. 

    \item \textbf{Edge resource constraints:} We consider the limited computation resource and storage capacity in Edge devices. PILOTE is designed to be with low computation and storage costs.
    \item \textbf{Extensive experiments on real-life data} : We apply PILOTE on real-life human activity data, which is collected by volunteers in our data collection campaigns.
    
\end{itemize}

The rest of this paper is organized as follows: Section~\ref{sec:related_work} presents the most related work to our proposal.
Section~\ref{sec:magneto_platform} shows the MAGNETO platform, where we deploy our Edge learning proposal for real-life human activity learning.
Section~\ref{sec:problemdef} introduces the problem definition. Section~\ref{sec:approach} formalizes the proposed method. An extensive set of experiments is performed and discussed in Section~\ref{sec:experiments}. We conclude and show future work in Section~\ref{sec:conclusion}. 
\section{Related work}\label{sec:related_work}
%JZ: newly added section
In this section, we show the most related work of our proposal in the Incremental Learning and Edge Machine Learning, with an application on Human Activity Recognition (HAR) task. 

\subsection{Incremental Learning}
% definition of the incremental learning
Incremental learning has been investigated for decades in the machine learning community. It refers to integrating new knowledge on the fly to the learned model, with unseen data coming in the future. 
% Scope of incremental learning  
In different contexts, the incremental learning is illustrated differently with various terminologies. It is also referred to as continual learning~\cite{de2021continual}, lifelong learning~\cite{liu2017lifelong} or sequential learning~\cite{aljundi2019selfless}. According to the learning targets, or the input and output distributions $P(\mathcal{X}^{(t)})$ and $P(\mathcal{Y}^{(t)})$ of an incoming batch data at time $t$, the incremental learning can be categorized into three types \cite{de2021continual}:
\begin{itemize}[leftmargin=*]
    \item \textit{Class incremental learning}: $\{\mathcal{Y}^{(t)} \subset \mathcal{Y}^{(t+1)}\}$ with $\{P(\mathcal{Y}^{(t)}) \neq P(\mathcal{Y}^{(t+1)})\}$
    \item \textit{Task incremental learning}: $\{\mathcal{Y}^{(t)} \neq \mathcal{Y}^{(t+1)}\}$
    \item \textit{Domain incremental learning}: $\{\mathcal{Y}^{(t)} = \mathcal{Y}^{(t+1)}\}$ with $\{P(\mathcal{Y}^{(t)}) = P(\mathcal{Y}^{(t+1)})\}$
\end{itemize}

% to constrain/clarify our research focus
Considering the newly emerged activities (i.e., classes) collected by Edge devices, we focus our attention in this paper on the \textit{class incremental learning} task.
% the shortcomings/missing points of the related work in class-incremental learning
A handful of works~\cite{li2017learning,rebuffi2017icarl,shin2017continual,kirkpatrick2017overcoming,prabhu2020gdumb} are devoted to handling the catastrophic forgetting problem in the \textit{class incremental learning} context. For instance, as a regularization-based method, Learning without Forgetting (LwF)~\cite{li2017learning} learns new classes, meanwhile conserves the learned information via knowledge distillation from the old classes; iCaRL~\cite{rebuffi2017icarl} keeps a set of class prototypes (i.e., memory replay) for representing the old classes, which are combined with new-class samples to update the model. Similarly, GDumb~\cite{prabhu2020gdumb} optimized the class prototype selection via Greedy Search; Instead of building up a memory replay for old-class data, Generative Replay~\cite{shin2017continual} learns a generative model and a solver (i.e., classifier) for learned data, that allows modeling the past class distribution instead of conserving raw data instances.

However, to push the incremental learning paradigm to the Edge, the above-mentioned methods are hardly applicable due to their context-specific model structure for Cloud-based processing. For instance, though LwF~\cite{li2017learning} does not require caching any past data for building a new model, the model itself is huge enough to be transferred or executed in Edge devices.

\subsection{Edge Machine Learning (Edge ML)}
Edge Machine Learning (Edge ML) emerges as a new paradigm, owning to its unbeatable advantages on low latency and strong privacy guarantee. In general, Edge ML can be characterized into two categories: i) Inference on the Edge, ii) Training on the Edge. 

For model inference on the Edge, past studies tend to optimize the model's scale and quantize its weights to reduce resource costs. For instance, commonly existing in Deep Neural Networks (DNNs), the model's parameter redundancies can be investigated to compress/re-design the model, while ensuring a low loss in performance. Typical model optimization methods include parameter pruning~\cite{han2015learning}, low-rank factorization~\cite{denton2014exploiting}, knowledge distillation~\cite{hinton2015distilling}, etc. These methods can be applied to different DNNs or be composed to optimize a complex model's inference on the Edge. Another branch of work~\cite{zhao2018deepthings} proposes to divide the DNN models and perform distributed inference computation.

For model training on the Edge, the resource cost is much higher than that in inference. Basically, the Edge training can be either based on tiny models requiring minimal resources, or splitting the training task via distributed/federated learning~\cite{yang2019federated}.

In this paper, we investigate the lightweight models which can be applied on Edge devices with limited resources.

\subsection{Human Activity Recognition (HAR)}
As a classic research problem in Data Mining and Machine Learning, Human Activity Recognition (HAR) gathers enormous studies regarding the data format, pre-processing techniques, learning models, post-processing methods, and model deployments. Among various data sources of HAR tasks, we consider human physical activities collected from wearable sensors. These activity data are generally represented by Time Series, more precisely, Multivariate Time Series (MTS)~\cite{zuo2021smate} on multiple sensors. 

As a classification task, the HAR model can be designed differently regarding data resources and targeted applications. For instance, one can use handcrafted features to feed any general ML models for downstream tasks, which is easy-to-deploy and requires linear processing time. 
More advanced work has been proposed in the Time Series Classification domain, where researchers aim to build general ML models covering various application domains~\cite{zuo2021smate}, including HAR tasks. For instance, authors in~\cite{zuo2019exploring} extract interpretable Shapelet features from time series, and combine with a kNN classifier; The recent end-to-end models~\cite{zuo2021smate} have shown promising results on HAR tasks, that generally rely on automatic feature extraction and selection~\cite{abboud2021micro,el2021tell} with regard to specific learning tasks, e.g., classification or forecasting. 

However, these approaches aim to reach optimal model performance with less attention on the model's scale or complexity, which are essential in the Edge context regarding the extremely limited resources. In this paper, we consider the HAR task as an application and focus on the Edge model's incremental learning behavior. Therefore, we adopt a primary feature extractor that relies on handcrafted statistic features, requiring linear processing time. Nevertheless, more advanced feature extractors can be explored and integrated into our framework, by considering the Edge constraints. This is orthogonal to our work.

\section{MAGNETO platform}
\label{sec:magneto_platform}
\begin{figure}[ht]
    \centering
    \includegraphics[width=1\linewidth]{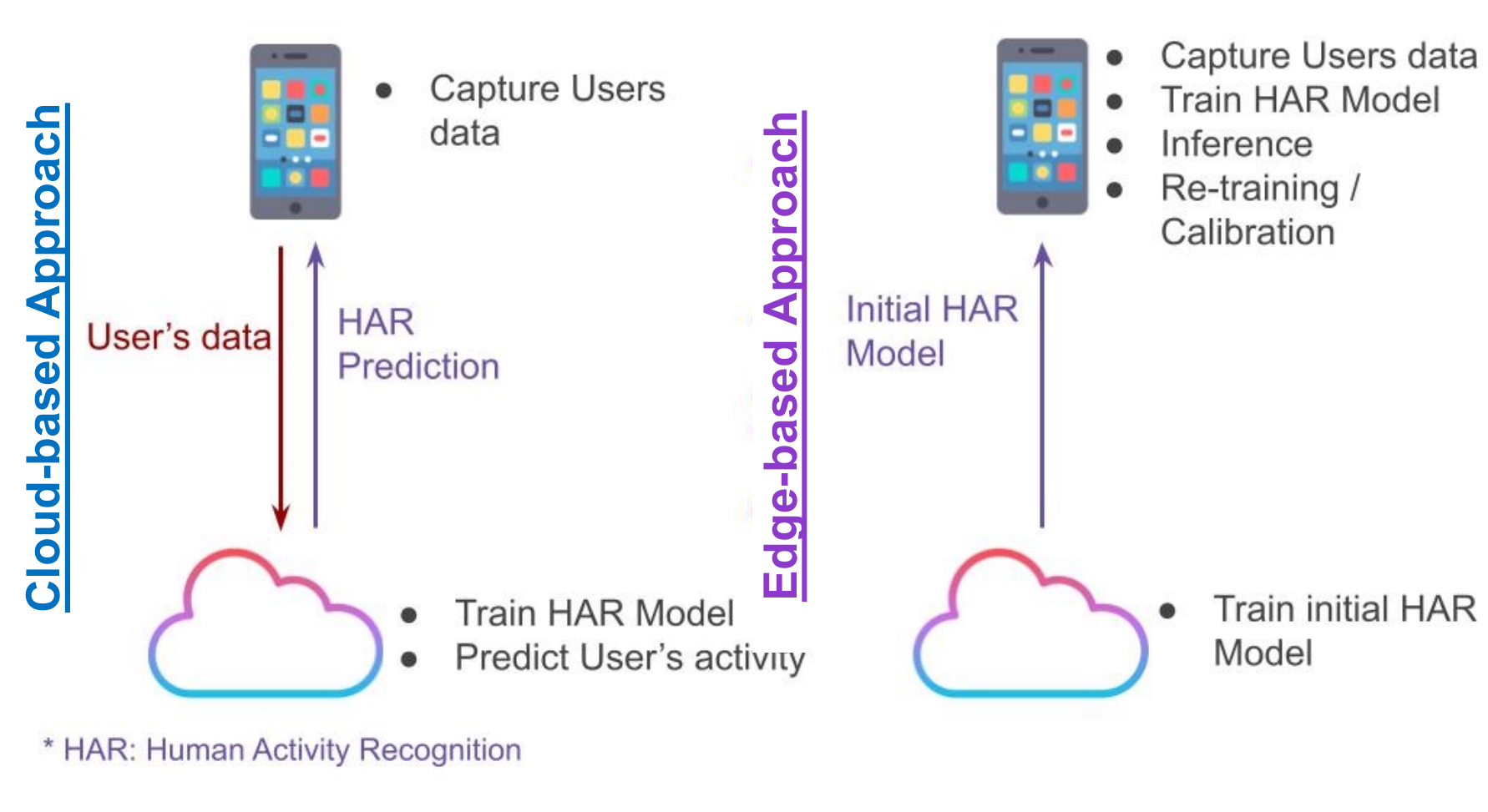}
    \caption{\textsf{MAGNETO processing steps:} (left side: Cloud-based architecture) (i) human activity data is captured on the \textsf{Edge} device, (ii) edge data is transferred to the Cloud, (iii) model training or inference on the Cloud, and (iv) send predictions back to the \textsf{Edge} device. (right side: \textsf{Edge-based} architecture) (i) an initial model is pretrained on the cloud, (ii) data is collected on the \textsf{Edge} device, (iii) the \textsf{Edge} device performs model's updating (with new data samples) or inference, (iv) the \textsf{Edge} device displays the predictions. %with Learning and inference happening on the Cloud and the activity estimation is transferred on the edge device, (Our Edge Approach: right side) Cloud transfer a pre-trained model on the Edge, and the inference as well as the Calibration and learning a new activity is happening on the Edge 
    }
    \label{fig:architecture_magneto}
\end{figure}

\textsf{MAGNETO}, \textit{sMArt sensinG for humaN activity rEcogniTiOn}, is a platform that provides inference on an Edge device without transferring user’s data to the Cloud. Figure~\ref{fig:architecture_magneto} shows its main architecture. The Cloud-based approach for Human Activity Recognition (HAR) tasks generally relies on continuous data exchange between Cloud and Edge, leading to high processing latency and privacy issues. \textsf{MAGNETO} is built primarily with an Edge-based structure, in which an initial HAR model is pre-trained on the Cloud as a warm starting point \cite{ash_warm-starting_2020} for the Edge learning process. The pre-trained model benefits from the rich computation resources on the Cloud computing center, meanwhile allowing further adaptation for new-coming real-time data on the Edge.

With limited computation resources but high flexibility, the Edge devices in MAGNETO platform provide the possibility for real-time data collection, model adaptation/re-training/calibration, and model inference. Importantly, all the operations are taking place on the Edge, and should not have any data exchange with the Cloud, considering data privacy issues. By adopting the MAGNETO platform, we are able to deploy PILOTE in a real industrial pipeline for Edge learning on real-time human activities.

\section{Problem formulation}
\label{sec:problemdef}

%In this section, we first formulate the incremental learning problem at the extreme edge. Then, we formally define the catastrophic forgetting problem when learning human activities at the edge. 

We consider a warm-start problem \cite{ash_warm-starting_2020} for learning new human activities on the edge: a learning model is pre-trained on the Cloud with its scalable resources, serving as a warm starting point for incremental learning on the edge. 
Table \ref{Math_notation} summarizes the notations used in the paper.

\begin{table}[htbp]
\centering
\caption{\label{Math_notation}Notation}
\scalebox{0.9}{
\begin{tabular}{|p{3cm}| l |}
\hline
\small
Notation & Description\\
\hline
\hline
$\mathcal{D}_o = (X^1, ..., X^{s-1})$ & Dataset of old classes $1, ..., s-1$\\
$\mathcal{D}_n = (X^s, ..., X^{t})$ & Dataset of new classes $s, ..., t$\\
$\mathcal{P} = (P^1, ..., P^{t})$ & Class exemplar sets\\
$\varphi: \mathcal{X} \to \mathcal{R}^d$ & Feature map function\\
$\Theta$ & Model parameters\\

\hline
\end{tabular}}
\end{table}

% distinguish new-class learning from incremental multi-task learning 
\begin{defn}(Incremental learning). Given a model $\Theta_o$ trained on $\mathcal{D}_o$, new sample set $\mathcal{D}_n = (X_{s}, ..., X_{t})$ with unknown activities $(s, ..., t)$ will enrich $\Theta_o$ incrementally, leading to a new model $\Theta_{n}$.
\end{defn}

\begin{defn}(Catastrophic forgetting). Given a newly updated model $\Theta_n$ trained sequentially on $\mathcal{D}_o$ and $\mathcal{D}_n$, the model tends to learn more from $\mathcal{D}_n$ but forget what has been already learned on $\mathcal{D}_o$, i.e., $\sum_{(x_i, y_i) \in \mathcal{D}_o}\mathcal{L}(f_{\Theta_n}(x_i), y_i)$ > $\sum_{(x_i, y_i) \in \mathcal{D}_o}\mathcal{L}(f_{\Theta_o}(x_i), y_i)$.
\end{defn}

Incremental learning of human physical activities on the edge requires considering not only the effect of catastrophic forgetting when updating the learning model, but also the extremely limited activity data collected by the edge devices, leading to imbalanced data distributions among classes. 
Therefore, given a model $\Theta_o$ trained on $\mathcal{D}_o$, a new sample set $\mathcal{D}_n$ comes with a small size $\left| \mathcal{D}_n \right| \ll \left| \mathcal{D}_o \right|$, the incremental learning at the extreme edge aims to learn a model $\Theta_n$, that guarantees the performance on $\mathcal{D}_n$:  $min(\sum_{(x_i, y_i) \in \mathcal{D}_o}\mathcal{L}(f_{\Theta_o}(x_i), y_i))$, and maintaining the performance on $\mathcal{D}_o$: $\sum_{(x_i, y_i) \in \mathcal{D}_o}\mathcal{L}(f_{\Theta_n}(x_i), y_i)\approx \sum_{(x_i, y_i) \in \mathcal{D}_o}\mathcal{L}(f_{\Theta_o}(x_i), y_i)$.

\section{Proposal: PILOTE}
\label{sec:approach}

\begin{figure*}[!htbp]
\centering
\includegraphics[width=1\linewidth]{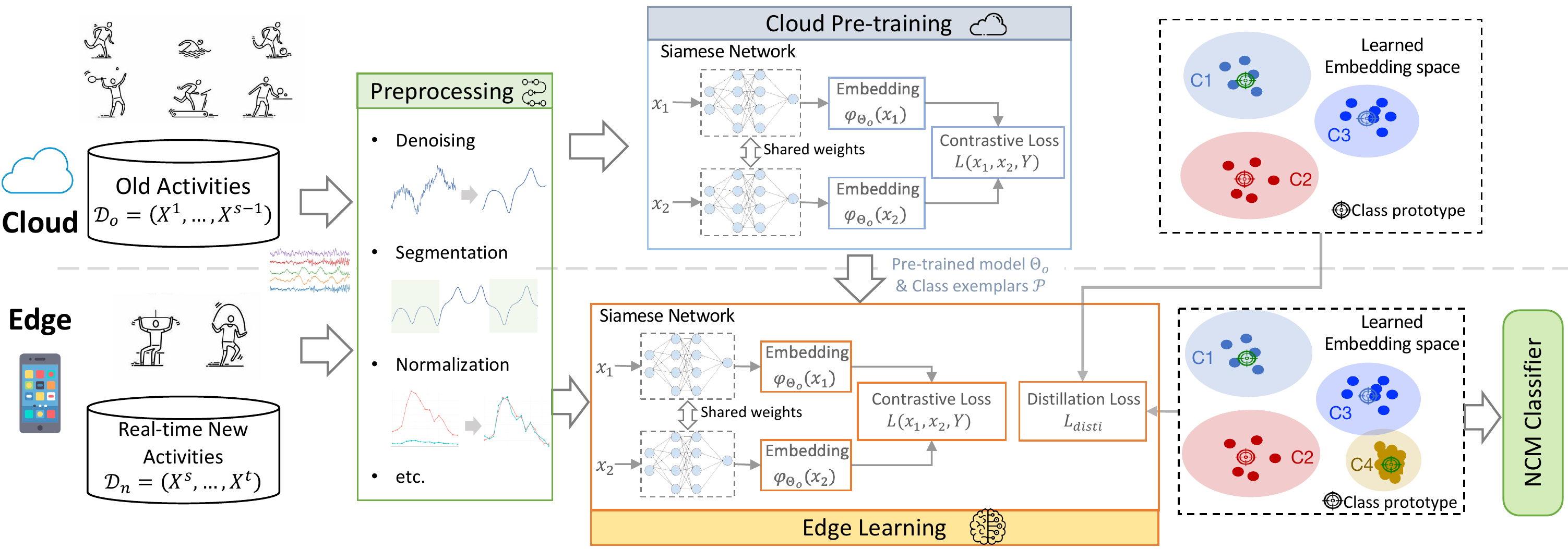}

\caption{Global system structure of PILOTE}
\label{system_structure}
\vspace{-1em}
\end{figure*}
%JZ: Add a global system structure showing the data collection, data preprocessing, cloud-edge interaction, embedding space building/evolving (catastrophic forgetting problem or class imbalance problem, etc.)

In order to answer the aforementioned challenges for incremental human activity learning on the edge, we propose PILOTE, a method that updates the learning model with new knowledge extracted from incoming new-class instances, while preserving the learned knowledge from existing classes.

% a brief presentation of the global structure of our learning model
Figure \ref{system_structure} shows the global system structure of PILOTE. The activity data collected from smartphone sensors are represented as Multivariate Time Series (MTS) \cite{zuo2021smate}. The preprocessing steps (e.g., denoising, segmentation, normalization, etc.), with linear time operations, are conducted equally on the Cloud and Edge devices. In particular, following the architecture design of the \textsc{MAGNETO} platform, PILOTE learns a feature representation from a data stream on the Edge, i.e., $X^1, ..., X^{t}$ in a class-incremental manner, where all examples of a set $X^y = \{x_{1}^{y}, ..., x_{n_y}^{y}\}$ are of class $y \in \mathcal{N}$. An incremental contrastive loss is designed to integrate new knowledge from $\mathcal{D}_n$, which forms the representation space gradually with a minimal computation cost. Meanwhile, inspired by knowledge distillation \cite{rebuffi_icarl_2017} for handling catastrophic forgetting problems, we propose a distillation loss which is designed explicitly for conserving the learned knowledge from existing datasets via guiding the representation space's building. Joint optimization of the distillation loss and contrastive loss allows for building the representation space in an extremely efficient manner. 

On the representation space, we apply the nearest class mean (NCM) classifier for a given sample $x_i$, which can be denoted as:
\begin{equation}
\begin{gathered}
y_i^*=\underset{y \in (1, ..., t)}{\arg \min } \operatorname{dist}\left(\varphi_{\Theta}(x_i), \mu^{y}\right), \\
\mu^{y}=\frac{1}{n_y} \sum_{p_{i}^{y}\in P^{y}}\varphi_{\Theta}(p_{i}^{y}),
\end{gathered}
\end{equation}
where $\mu^{y}$ denotes the prototype of class $y$, $P^y$ is the exemplar set of class $y$ and $n_y$ is the number of exemplars in class $y$.

\subsection{Incremental representation learning}
% Motivation for applying few-shot learning, e.g., Siamese Network
Compared to large volumes of data adopted for training the learning model on the cloud, new human activities recorded at edge devices usually show limited data samples. The imbalanced data distribution will lead to training constraints for learning the latest activities. To this end, instead of building an end-to-end classification model, we adopt the concept of Few-shot Learning: a Siamese Network \cite{koch2015siamese} is applied to learn a robust representation space, for which much less data is required. Instead of learning feature-label mappings, the Siamese Network only learns separable boundaries in the representation space between the activities. In this space, the intra-class embeddings are clustered, whereas the inter-class embeddings are clearly bounded. A set of class prototypes are generated to reduce the computation cost of the NCM classifier. The supervised contrastive loss \cite{khosla_supervised_2020} is applied to the representation space: 
\begin{equation}
\small
\label{eq:cont_loss}
\begin{aligned}
L\left(x_i, x_j, Y\right) &=Y *\left\|\varphi_{\Theta}\left(x_i\right)-\varphi_{\Theta}\left(x_j\right)\right\|^2 \\
&+ (1-Y) *\left\{\max \left(0, \textbf{m}^2-|| \varphi_{\Theta}\left(x_i\right)-\varphi_{\Theta}\left(x_j\right) \|^2\right)\right\}
\end{aligned}
\end{equation}
where $\textbf{m} >0$ is the margin parameter, $Y=1$ if $x_i, x_j$ are similar (i.e., $y_i=y_j$), otherwise $Y=0$.

\begin{algorithm}[htbp]
%\SetAlCapNameFnt{\scriptsize}
%\SetAlCapFnt{\scriptsize}
%\scriptsize
  \KwIn{{$\Theta_{o}$} \tcc*{Current model parameters}} 
  \KwIn{K \tcc*{Edge cache size} }
  \KwIn{{$\mathcal{D}_o = (X^1, ..., X^{s-1})$} \tcc*{Old sample sets of classes $1, ..., s-1$}} 
  \KwIn{{$\mathcal{D}_n = (X^s, ..., X^{t})$} \tcc*{New sample sets of classes $s, ..., t$}}

  \KwOut{{$\Theta_{n}$} \tcc*{Updated model parameters}} 
  \KwOut{$\mathcal{P} = (P^1, ..., P^{t})$ \tcc*{Exemplar support sets}} %$offset$ 

    \tcc{(Cloud) Exemplars selection from old-class samples}
    $m = K/(s-1) $\\
    
    \For{$y \in (1, ..., s-1)$}{
        $X^y = (x_{1}^y, ..., x_{n_y}^{y})$ \\
        
        $\mu^{y} =\frac{1}{n_y} \sum_{i=1}^{n_y} \varphi_{\Theta_o}\left(x_i\right)$ \\

        \tcc{Select m closest samples in the embedding space}
        \For{$k \in (1, ..., m)$}{
            $p_{k}^{y} \leftarrow \underset{x \in X^y}{\operatorname{argmin}}\left\|\mu^{y} - \frac{1}{k}\left(\varphi_{\Theta_o}(x)+\sum_{j=1}^{k-1} \varphi_{\Theta_o}(p_{j}^{y})\right)\right\|$
        }

        $P^{y} \leftarrow (p_{1}^{y}, ..., p_{m}^{y})$
    }
    
    \tcc{(Edge) Incremental Representation Learning with new-class samples}
    $D_0 \leftarrow \bigcup_{y=1, \cdots, s-1}\left\{(x, y): x \in P^y\right\}$ \\

    $D_n \leftarrow \bigcup_{y=s, \cdots, t}\left\{(x, y): x \in X^{y}\right\}$\\

    Run network training with an updated parameter set $\Theta_{n}$ with loss function $L = \alpha \times L_{\text{disti}} + (1-\alpha) \times L_{\text{contra}}$, where \\
     
    $L_{\text{disti}} =\sum_{\left(x_i, \_\right) \in D_0}\left\| \varphi_{\Theta_n}\left(x_i\right)-\varphi_{\Theta_o}\left(x_i\right)\right\|^{2} $\\
  
    $L_{\text{contr}} =\sum_{\left(x_i, y_i\right) \in D_0} \sum_{\left(x_i, y_i\right) \in D_n} L\left(x_i, x_j, Y \leftarrow y_i == y_j\right)$
  
  \Return  $\Theta_{n}$
  \caption{Incremental Representation Learning considering Catastrophic Forgetting on the Edge}
  \label{IncrementalRepresentationLearning}

\end{algorithm}

As shown in Algo. \ref{IncrementalRepresentationLearning}, we first select the most representative exemplars from old-class samples as a support set, referring to the $m$ closest samples to each class prototype $\mu^{y}$. The support set will be transferred along with a pre-trained model $\Theta_{o}$ from the Cloud to the Edge devices. Then, the representation space is built over the support set and new-class samples. The exemplar sets and class prototypes are updated continuously with newly integrated samples. 
Following this process, the approach incrementally learns new classes. However, it suffers from the limitation of forgetting past knowledge. Therefore, to complement this method, we investigate the problem and propose a solution, discussed in the next section. 
% Time complexity of the incremental contrastive/triplet loss
% for instance, the number of pairs in Contrastive Loss can be reduced to n_old * n_new, instead of (n_old + n_new) ** 2, as the representations learned from old instances are already constrained by Distillation Loss.

\subsection{Catastrophic forgetting at the extreme edge}
%catastrophic forgetting problem in the representation space
When learning from new activities, the model tends to forget what has been learned from previous activities. Precisely, the boundary between existing activities in the embedding space becomes unclear after learning new activities. Therefore, it is required to maintain the learned boundary between the old-class samples, while learning new activity embeddings by staying far away from existing ones.

As a concept from Transfer Learning, Distillation Loss \cite{hinton2015distilling} has been widely adopted for extracting key knowledge from a trained neural network. We apply the Distillation Loss under the few-shot learning context, where the previously learned model can be constrained while integrating newly learned knowledge from extremely limited instances. 
Importantly, the introduced constraints allow simplifying the computations of the Contrastive Loss, thus reducing the time complexity of the incremental learning process. For instance, given $n_t$ new samples of class $t$, the number of pairs in Contrastive Loss can be reduced to $\frac{n_{t}!}{2! (n_{t} - 2)!}$, instead of $\sum_{y \in (1, ..., t)}t*\frac{n_{y}!}{2! (n_{y} - 2)!}$, where $n_y$ is the number of samples in class $y$. In the embedding space, the learned representations from old-class instances are already constrained by Distillation Loss.

Finally, by combining both the Distillation Loss from old-class instances and the Contrastive Loss of new-class instances with a balancing weight $\alpha$, we are capable of learning a robust incremental learning model with constant memory.
\section{Experiments}
\label{sec:experiments}
In this section, we demonstrate the effectiveness of PILOTE with real-life human activity datasets. The experiments were designed to answer the following questions:

\begin{itemize}[leftmargin=.2in]
    
    \item[Q1] \textit{Catastrophic forgetting:} How successful is our model at learning new activities while considering the catastrophic forgetting problem? % compare models' performance with or without considering catastrophic forgetting

    % why our method can be applied on the edge: regarding the size of the support set; training speed; show the impact of sampling strategy
    \item[Q2] \textit{Applicability on the edge:} How well does our model perform on the edge with limited storage and computing resources?
    
    \item[Q3] \textit{Extreme edge with minimal new-class samples:} How does our model perform when new-class samples are extremely limited? % size of new-class samples < old-class samples in support set

\end{itemize}

\subsection{Experimental settings}
\subsubsection{Dataset description}
%JZ: Add more details about the data collection campain, e.g., how many participants, collection devices, activities, campain/activity durations, preprocessing techniques, etc.  
We base our experiments on real human physical activity data collected on edge devices. 
We have launched data collection campaigns, capturing an initial dataset of more than 100GB of sensory data. Even though the preprocessing was conducted on the cloud, the real-time coming data can be processed instantly, as the preprocessing operation requires linear time. Precisely, we split the sensory data into a one-second recording window with roughly 120 sequential measurements from 22 mobile sensors, e.g., accelerometer, gyroscope, and magnetometer. We extract 80 statistical features such as the average, the variance for each feature, the average jerk, and the variance of the jerk for each three-dimensional feature sensor. Currently, five activities with $\sim 200k$ %199839
records are collected: `Drive', `E-scooter', `Run', `Still', `Walk'. To simplify the test scenarios and validate the learning model, we design the incremental learning process by setting one of the activities as the new-class data to be learned.

\subsubsection{Execution and Parameter Settings} The proposed model is implemented in PyTorch 1.6.0 and is trained using the Adam optimizer. We set an adaptive learning rate regarding training epochs, i.e., the learning rate starts from 0.01 and decreases by half every training epoch. The backbone model is a simple Fully Connected (FC) neural network with dimensions [1024 x 512 × 128 × 64 × 128]. We apply Batch Normalization \cite{IoffeBatchShift} and Rectified Linear Units (ReLU) \cite{NairRectifiedMachines} activation functions on the first four layers. The last layer  projects the input samples into a representation space with an embedding size of 128. The backbone model can be any other advanced network structure. Here we focus on the model's incremental learning behaviors. We set the balancing weight $\alpha=0.5$ in all experiments. We split 30\% of the collected records as the test set. The validation split is set to 0.2 for both pre-training and incremental training. We set the stopping condition to hold when the difference of validation loss between epochs is less than a small threshold, 0.0001 for five consecutive steps. We execute each model in five rounds and report the average accuracy and the standard deviations.

\subsubsection{Baselines} 
To the best of our knowledge, previous work rarely considers the extremely limited learning data and catastrophic forgetting problem simultaneously in the Edge learning context. Therefore, we compare PILOTE with two popular strategies for human activity recognition on the edge:

\begin{itemize}[leftmargin=.2in]
    
    \item[1] \textit{Pre-trained model:} The model is pre-trained on the cloud on four activities. It is transferred to the edge with a support set. The model generates class prototypes for new-class samples and enriches the support set with random new-class data. 

    \item[2] \textit{Re-trained model:} The pre-trained model is re-trained on the edge using the enriched support set with new-class samples.
    
\end{itemize}

\subsection{Q1: Catastrophic forgetting}
% check the results for every missing class in the label set
% the preliminary results are obtained from one single missing/new class 
To validate the model's robustness in handling the catastrophic forgetting problem, we pick one of the five activities each time as the new class to be learned. Table \ref{results_catastrophic_forgetting} shows the models' performance comparisons on the five new-class scenarios. The re-trained model and PILOTE in each scenario are based on the same pre-trained model. We report the average accuracy and the standard deviations of five executions for both the re-trained model and PILOTE.  

\begin{table}[!htbp]
\centering
\caption{Accuracy comparison of learning models without and with considering the catastrophic forgetting problem.}
\label{results_catastrophic_forgetting}
\begin{tabular}{cccc} 
\toprule
New class & Pre-trained & Re-trained   & PILOTE                       \\ 
\midrule
Drive         & 0.8443                 &0.8825$\pm$0.0391	& \textbf{0.8837$\pm$0.0278}                                                \\
E-scooter     & 0.8744                 & 0.9491$\pm$0.0089	&  \textbf{0.9516$\pm$0.0074}           \\
Run           & 0.7856                 & 0.9126$\pm$0.0321 & \textbf{0.9372$\pm$0.0319}  \\
Still         & 0.8213                 &0.9143$\pm$0.0288	&\textbf{0.9349$\pm$0.0340}  \\
Walk          & 0.8376           &0.8909$\pm$0.0383     &\textbf{0.9193$\pm$0.0386}                             \\
\bottomrule
\end{tabular}
\end{table}

\begin{figure}[!htbp]
\centering 
\subfloat[Re-trained model]{
    \includegraphics[width=1\linewidth]{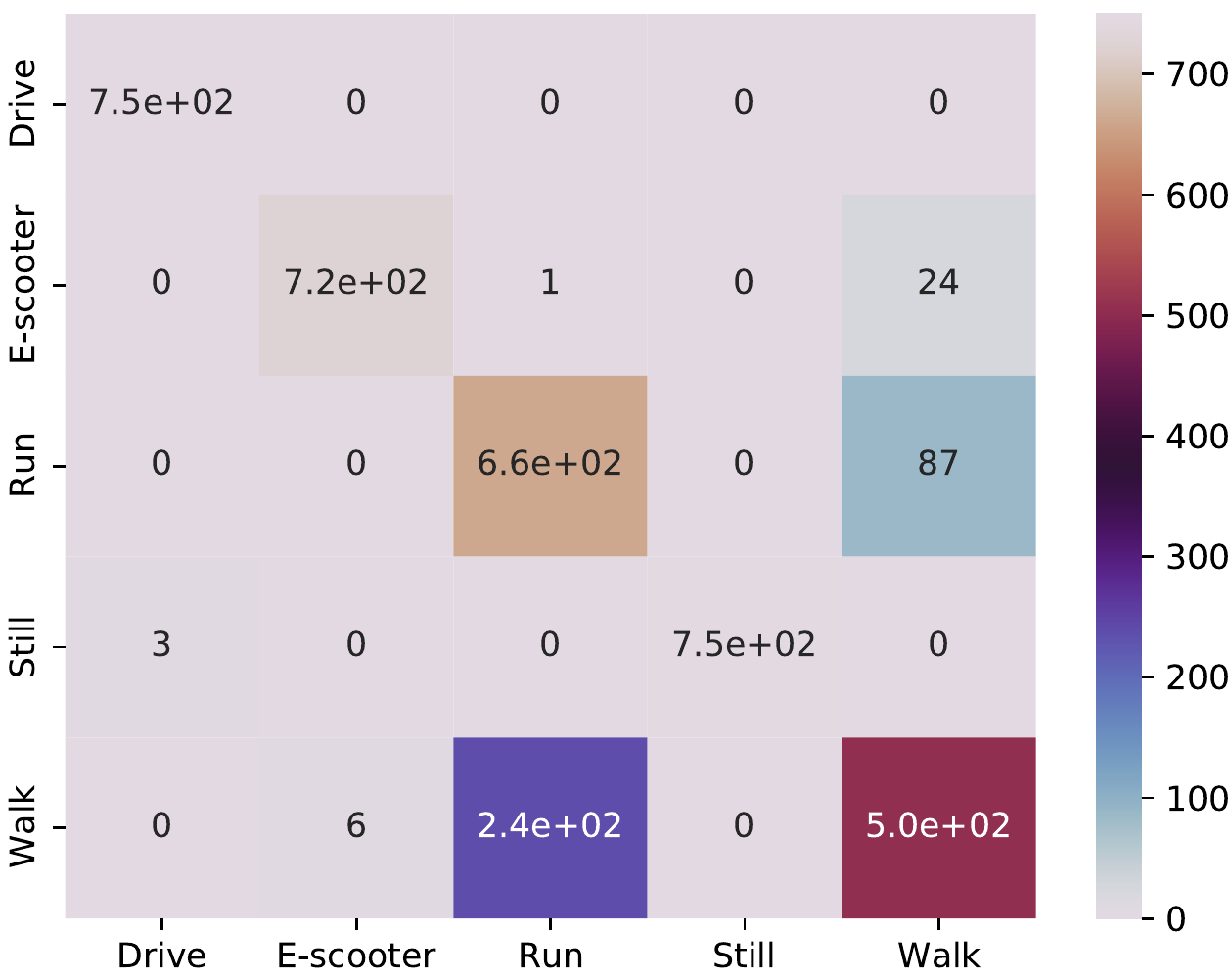}
}
\newline
\centering
\subfloat[PILOTE]{
    \includegraphics[width=1\linewidth]{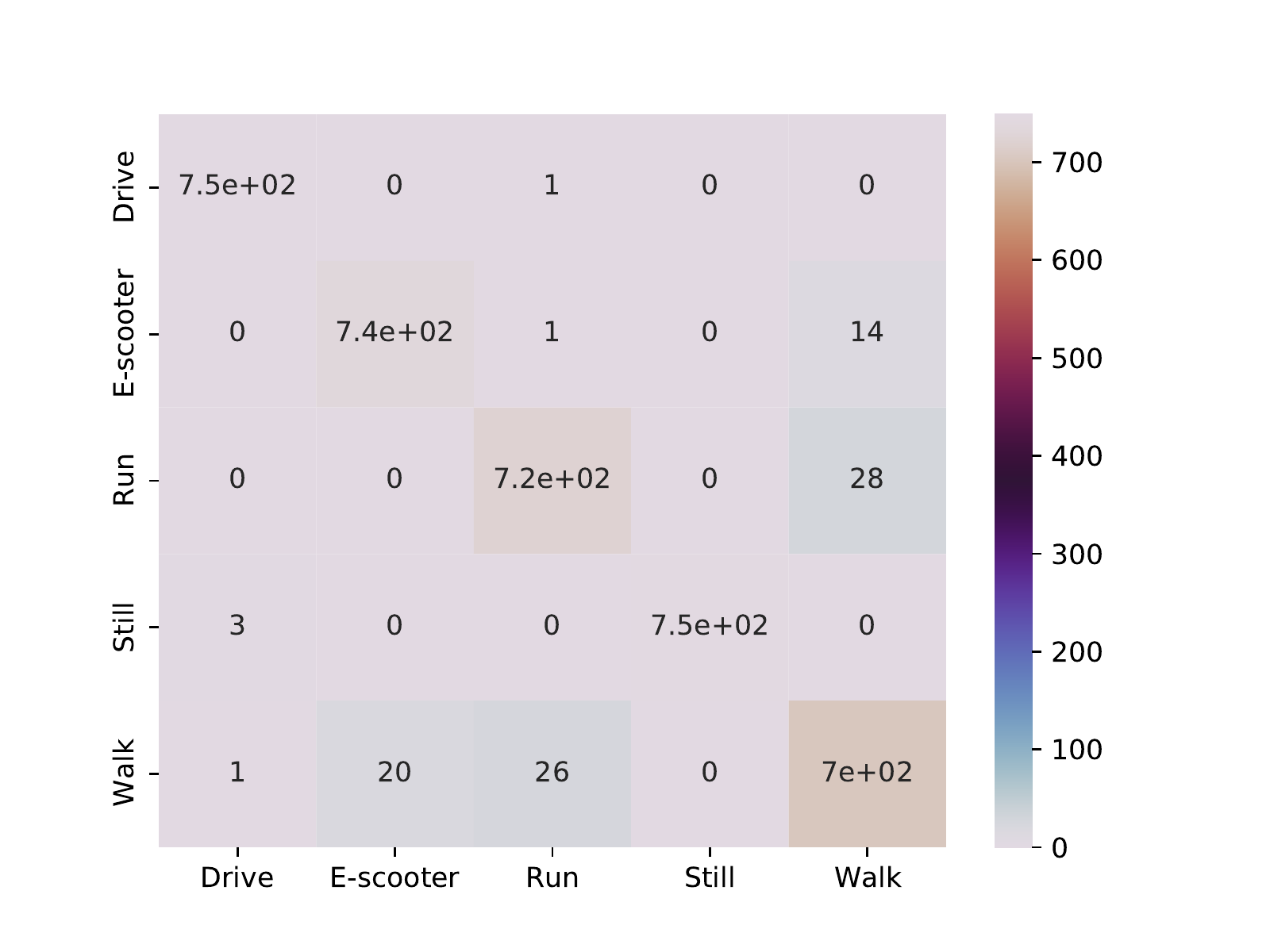}
}
\centering 
\hfill
\caption{Confusion matrices for models trained on the new class `Run', with 200 exemplars per class in the support set.}
\label{conf_matrix_catastrophic_forgetting}
\vspace{-1em}
\end{figure}

\begin{figure*}[!htbp]
\centering 
\subfloat[Pre-trained model]{
    \includegraphics[width=0.33\linewidth]{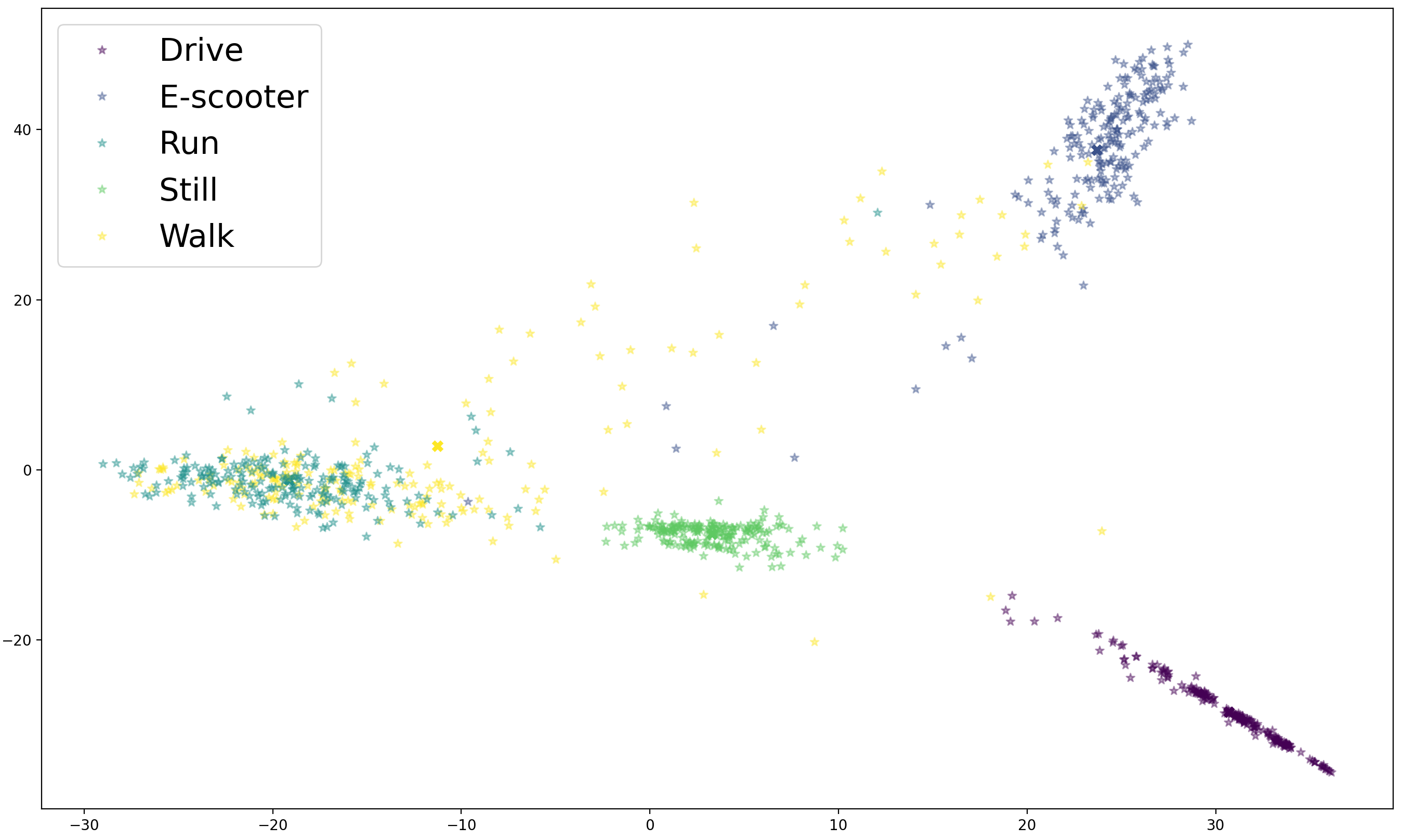}
}
\centering
\subfloat[Re-trained model]{
    \includegraphics[width=0.33\linewidth]{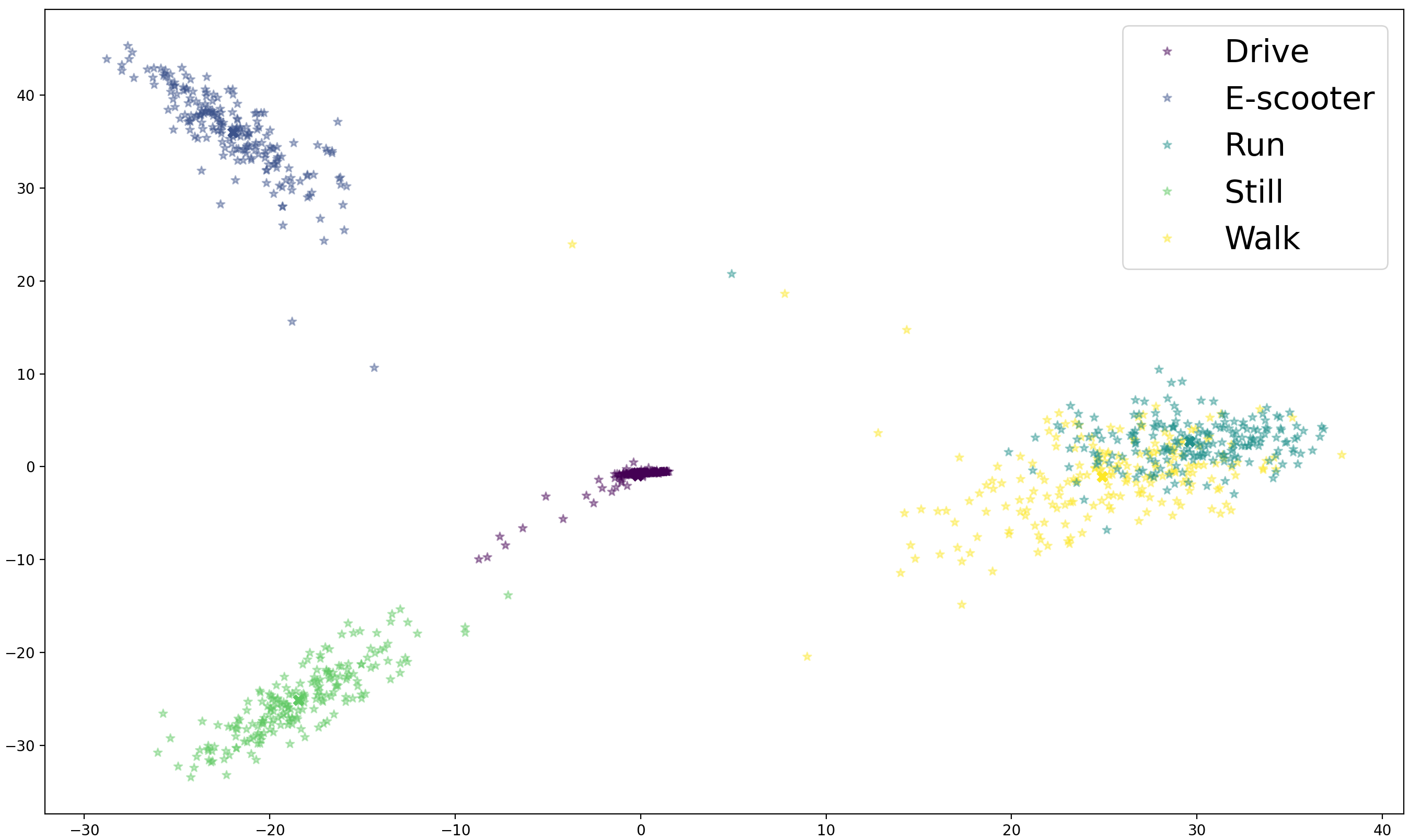}
}
\centering
\subfloat[PILOTE]{
    \includegraphics[width=0.33\linewidth]{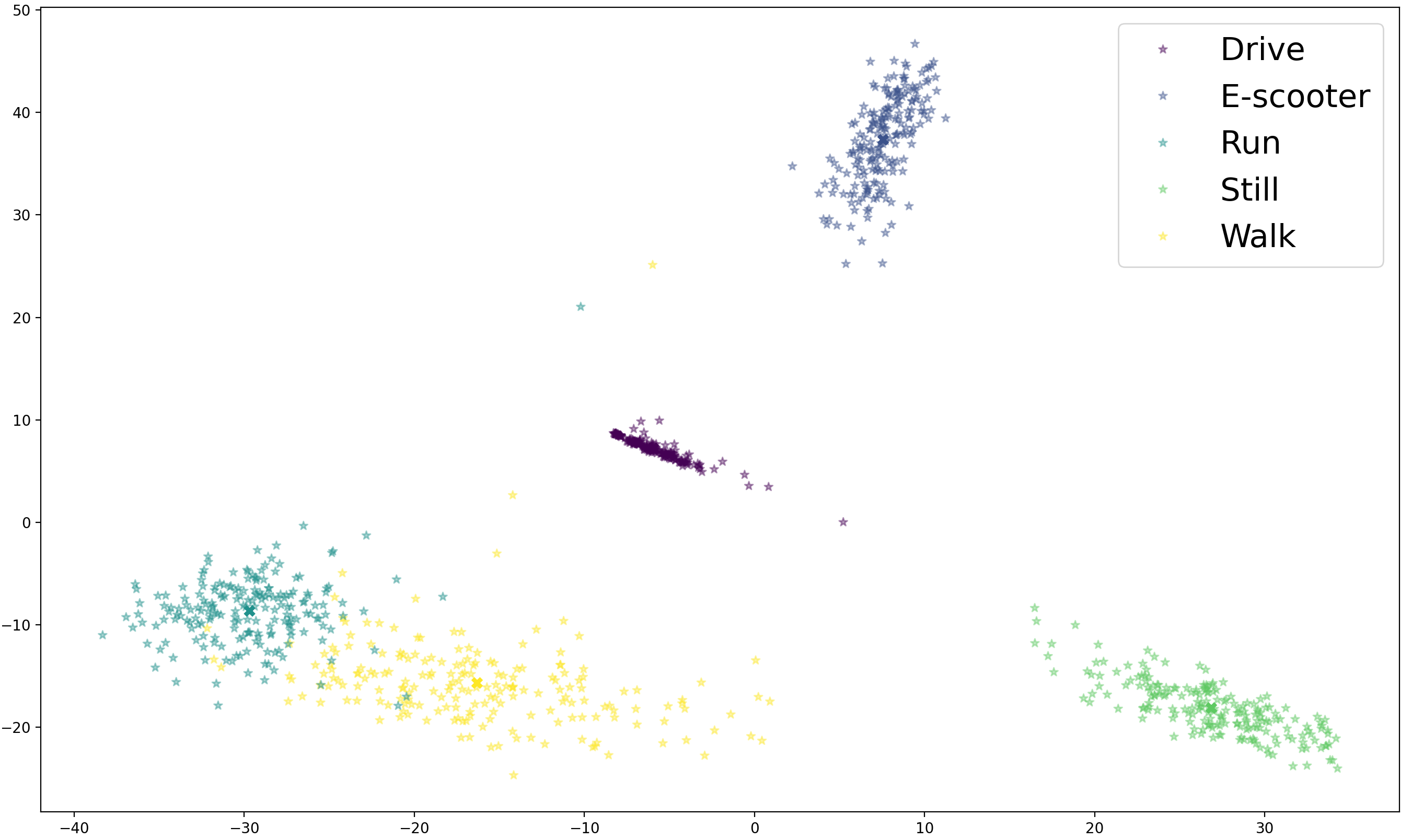}

}
\centering 
\hfill
\caption{Visualization of the embedding spaces of the learning models. The `Run' activity is excluded from the pre-training process. The support set includes 200 representative exemplars for each class.}
\label{visualization_embeddings}
\vspace{-1em}
\end{figure*}

Table \ref{results_catastrophic_forgetting} suggests that PILOTE does achieve better performances by conserving the learned knowledge from old-class samples. In particular, for activities `Run', `Still', and `Walk', PILOTE obtained more than 2\% performance improvement over the re-trained model. These activities are more challenging to learn and distinguish than `Drive' and `E-scooter'. This can also be illustrated by Figure \ref{conf_matrix_catastrophic_forgetting}, which shows confusion matrices when learning the new activity `Run'. 
The confusion matrices suggest that the re-trained model tends to learn from `Run' but forgets what has been learned from `Walk', leading to a massive amount of false positives when predicting `Run', and false negatives for `Walk'. Whereas PILOTE allows separating the two similar activities by conserving the knowledge learned previously from `Walk'. Figure \ref{visualization_embeddings} shows the visualization of the embedding spaces of the three models, which further validates our claims: a re-trained model can better separate `Run' and `Walk' than a pre-trained model, but shows a more blurred boundary than PILOTE.

\subsection{Q2: Applicability on the edge}
\begin{figure}[!htbp]
\centering
\includegraphics[width=0.9\linewidth]{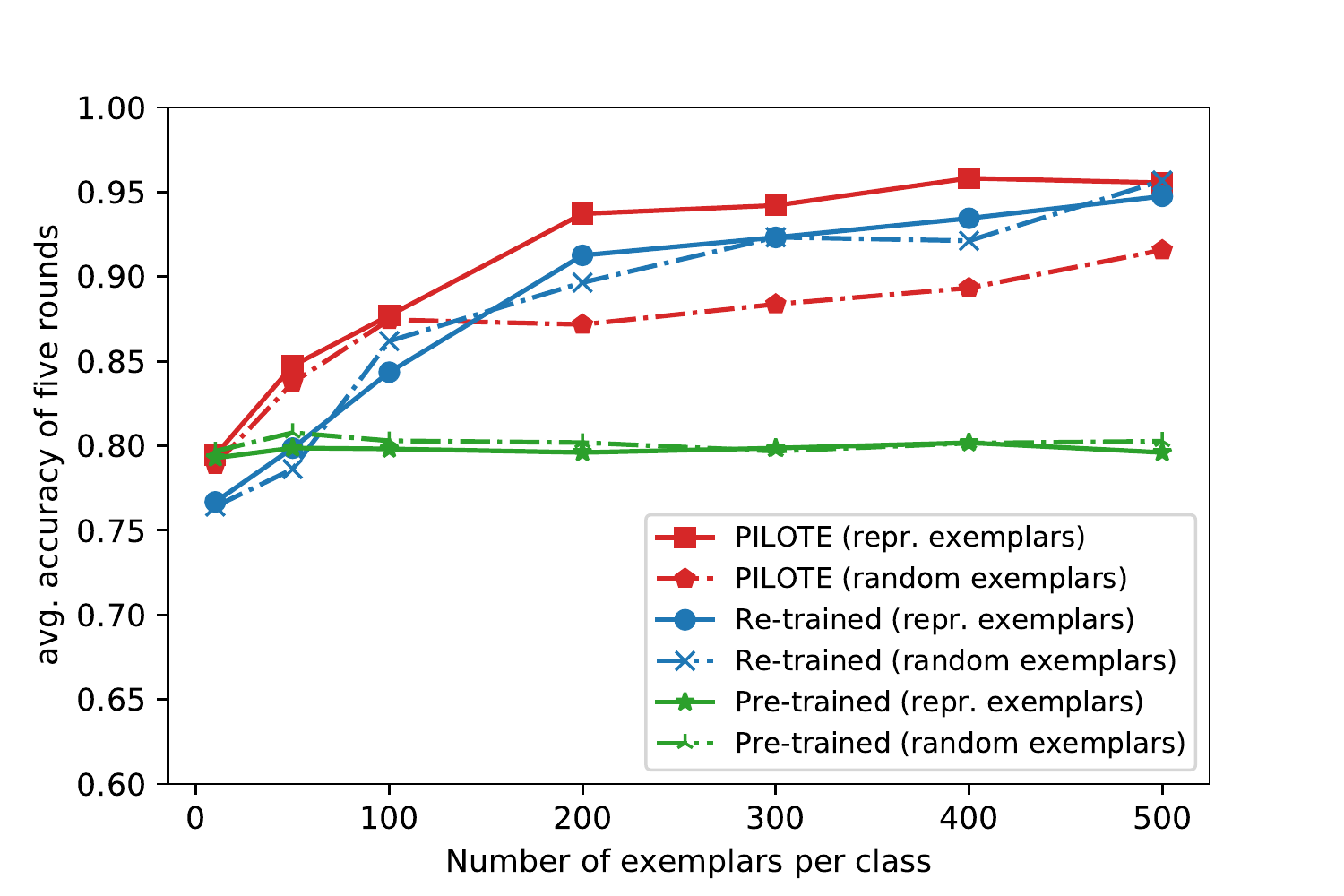}
\caption{Model accuracy regarding the support set's size.  The `Run’ activity is excluded from the pre-training process}
\label{p2_performance_support_size}
\end{figure}

% 1. to show the model's performance regarding the size of the support set
% 2. to show the impact of different sampling strategies
% -> Four models regarding the size of the support set: a) PILOTE with exemplar selection; b) PILOTE without exemplar selection; c) Re-training with exemplar selection; d) Re-training without exemplar selection
With limited computing and storage resources on the edge, the learning process is required to be efficient, and effective when learning from a small amount of data. In Figure \ref{p2_performance_support_size}, we validate the models' performance regarding the support set's size or the edge devices' storage space, e.g., 2500 exemplars in compressed format would take 3.2MB of space. We show in Figure \ref{p2_performance_support_size} as well the impact on learning models of different exemplar selection strategies.

From Figure \ref{p2_performance_support_size}, the learning models tend to have better performance with more class exemplars. Except for the pre-trained models, the performance stays stable regarding the number of exemplars, as the old-class embeddings are clearly separated. The representative exemplars impact PILOTE more than other models, especially for an extensive support set with more than 200 exemplars per class. This is probably caused by the deviation between the randomly selected exemplars and class prototypes, affecting PILOTE's knowledge distillation. The re-trained model rebuilds the embedding space, showing less dependence on the selected exemplars.
We can observe as well from Figure \ref{p2_performance_support_size} that with a few exemplars (e.g., < 50), the re-trained models perform even worse than the pre-trained models, whereas PILOTE performs better. This is mainly due to over-fitting the small amount of data while forgetting the learned knowledge from old-class samples. When adding learning constraints, PILOTE can mitigate the impact of catastrophic forgetting and learn from a small sample set. With less than 200 exemplars per class (i.e., < 256KB), PILOTE can reach an accuracy of 93.72\% within 20 training epochs, and each epoch costs less than 0.5s. These features make PILOTE highly suitable to be deployed on edge devices.

\subsection{Q3: Extreme edge with a few samples}
% baseline can be non-siamese network-based methods (leave it for future work) 
% here we just show the model's performance with a few new-class samples (with PILOTE and retrained models). we fix the support set size for old-class samples and vary the size of new-class samples to check the model's performance. 
In practice, new activities are recorded on edge devices, the data volume is not as large as that collected via a centralized manner. Very limited data samples can be recorded at the extreme edge for learning new activities, leading to an unbalanced data distribution among classes. To validate the model's performance at the extreme edge, we vary the amount of new-class samples (i.e., `Run') in the support set, and select 200 representative exemplars for old-class samples. We need to note that the new-class exemplars are randomly selected from new-class samples.

\begin{figure}[!htbp]
\centering
\includegraphics[width=0.9\linewidth]{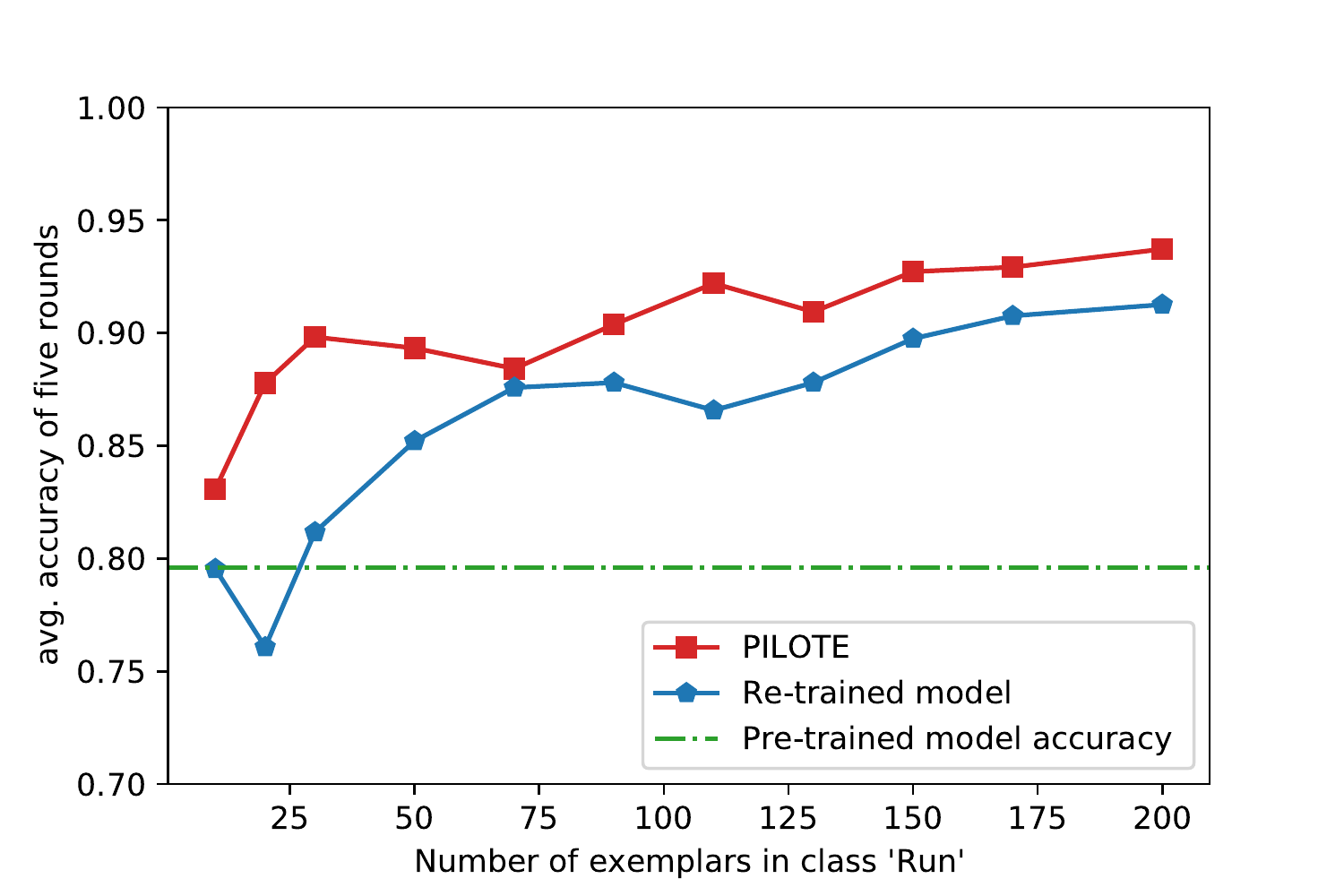}
\caption{Model accuracy regarding the exemplar number in `Run'. Each class includes 200 exemplars in the support set.}
\label{p3_performance_support_size}
\vspace{-1em}
\end{figure}

Figure \ref{p3_performance_support_size} shows the performance comparison between PILOTE and the re-trained model regarding the number of new-class exemplars in the support set. We also show the pre-trained model's accuracy as a warm starting point \cite{ash_warm-starting_2020}. The results suggest that only with 30 exemplars in class `Run', PILOTE can obtain a 90\% accuracy score. Globally, PILOTE performs better than the re-trained model, particularly when the exemplar number is small (e.g., < 50). When the previously learned knowledge is conserved, the model requires less effort/data for updating the embedding space, thus becoming more robust when new-class samples are extremely limited.

\section{Conclusion and perspectives}
\label{sec:conclusion}
% conclusion
In this paper, we proposed PILOTE, an incremental learning model applicable to the extreme edge, which considers both the limited edge resources and the catastrophic forgetting problem.
% perspectives: more complex data with more activities  
In future work, we aim to launch more campaigns for collecting more complex data with more human physical activities. The enriched data would help validate PILOTE in more practical and challenging scenarios. From a longer-term perspective, one can consider the model's scaling up or collaborative learning with strong privacy-preserving guarantees, e.g., Federated Learning.
%\begin{acks}
%To Robert, for the bagels and explaining CMYK and color spaces.
%\end{acks}

\bibliographystyle{ACM-Reference-Format}
\bibliography{references}

\end{document}